\documentclass{article}

\usepackage{arxiv}

\usepackage[utf8]{inputenc} 
\usepackage[T1]{fontenc}    
\usepackage{hyperref}       
\usepackage{url}            
\usepackage{booktabs}       
\usepackage{amsfonts}       
\usepackage{nicefrac}       
\usepackage{microtype}      
\usepackage{lipsum}		
\usepackage{graphicx}
\usepackage{natbib}
\usepackage{doi}
\usepackage{hyperref}
\usepackage{natbib} 
\usepackage{mathtools} 
\usepackage{amssymb} 
\usepackage{xcolor} 
\usepackage{siunitx} 
\usepackage{booktabs} 
\usepackage{tikz} 

\renewcommand{\undertitle}{}
\renewcommand{\headeright}{}
\date{}
\title{ALIGN: Adversarial Learning for Generalizable Speech Neuroprosthesis}

\author{
Zhanqi Zhang \thanks{These authors contributed equally.}  \\
Department of Computer Science and Engineering\\
University of California San Diego\\
La Jolla, California, USA \\
\texttt{zhanqi@ucsd.edu} \\
\And
Shun Li \footnotemark[1]  \\
Kempner Institute for the Study of \\ Natural and Artificial Intelligence\\
Harvard University\\
Boston, Massachusetts, USA \\
\texttt{shunli@g.harvard.edu} \\
\And
Bernardo L. Sabatini \\
Kempner Institute for the Study of \\ Natural and Artificial Intelligence\\
Harvard University\\
Boston, Massachusetts, USA \\
\texttt{bernardo\_sabatini@hms.harvard.edu} \\
\And
Mikio Aoi \thanks{These authors contributed equally.} \\
Halıcıoğlu Data Science Institute\\
Department of Neurobiology\\
University of California San Diego\\
La Jolla, California, USA \\
\texttt{maoi@ucsd.edu} \\
\And
Gal Mishne \footnotemark[2]  \\
Halıcıoğlu Data Science Institute\\
Department of Electrical and Computer Engineering\\
Department of Computer Science and Engineering\\
University of California San Diego\\
La Jolla, California, USA \\
\texttt{gmishne@ucsd.edu} \\
}

\renewcommand{\shorttitle}{Preprint}

\hypersetup{
pdftitle={A template for the arxiv style},
pdfsubject={q-bio.NC, q-bio.QM},
pdfauthor={David S.~Hippocampus, Elias D.~Striatum},
pdfkeywords={First keyword, Second keyword, More},
}

\begin{document}
\maketitle

\begin{abstract}
Intracortical brain-computer interfaces (BCIs) can decode speech from neural activity with high accuracy when trained on data pooled across recording sessions. In realistic deployment, however, models must generalize to new sessions without labeled data, and performance often degrades due to cross-session nonstationarities (e.g., electrode shifts, neural turnover, and changes in user strategy). In this paper, we propose \textbf{ALIGN}, a session-invariant learning framework based on multi-domain adversarial neural networks for semi-supervised cross-session adaptation. ALIGN trains a feature encoder jointly with a phoneme classifier and a domain classifier operating on the latent representation. Through adversarial optimization, the encoder is encouraged to preserve task-relevant information while suppressing session-specific cues. We evaluate ALIGN on intracortical speech decoding and find that it generalizes consistently better to previously unseen sessions, improving both phoneme error rate and word error rate relative to baselines. These results indicate that adversarial domain alignment is an effective approach for mitigating session-level distribution shift and enabling robust longitudinal BCI decoding.
\end{abstract}

\section{Introduction}\label{sec:intro}

Intracortical brain–computer interfaces (BCIs) have recently achieved strong brain-to-text performance (\cite{willet}), but a central challenge for real-world use is cross-session generalization. Neural recordings are nonstationary: the mapping from neural activity to intended output drifts across sessions due to electrode impedance changes, electrode drift, and neural turnover (\cite{nonstationary, nonsta2, nonsta3}). As a result, decoders trained on one session can degrade on later sessions, requiring frequent recalibration and limiting long-term usability. For participants, each additional day of calibration is a substantial burden, adding clinical workload and reducing time available for everyday communication. Moreover, it is well known in neuroscience that biological systems naturally execute motor programs like vocalization/movement at diverse speeds with high temporal flexibility (\cite{songbird, wang2018flexibletiming, gainmodulation2018}). Therefore, developing methods to mitigate such trial- and session-level variability is critical.

\begin{figure*}[!htb]
  \centering
  \includegraphics[width=1.0 \linewidth]{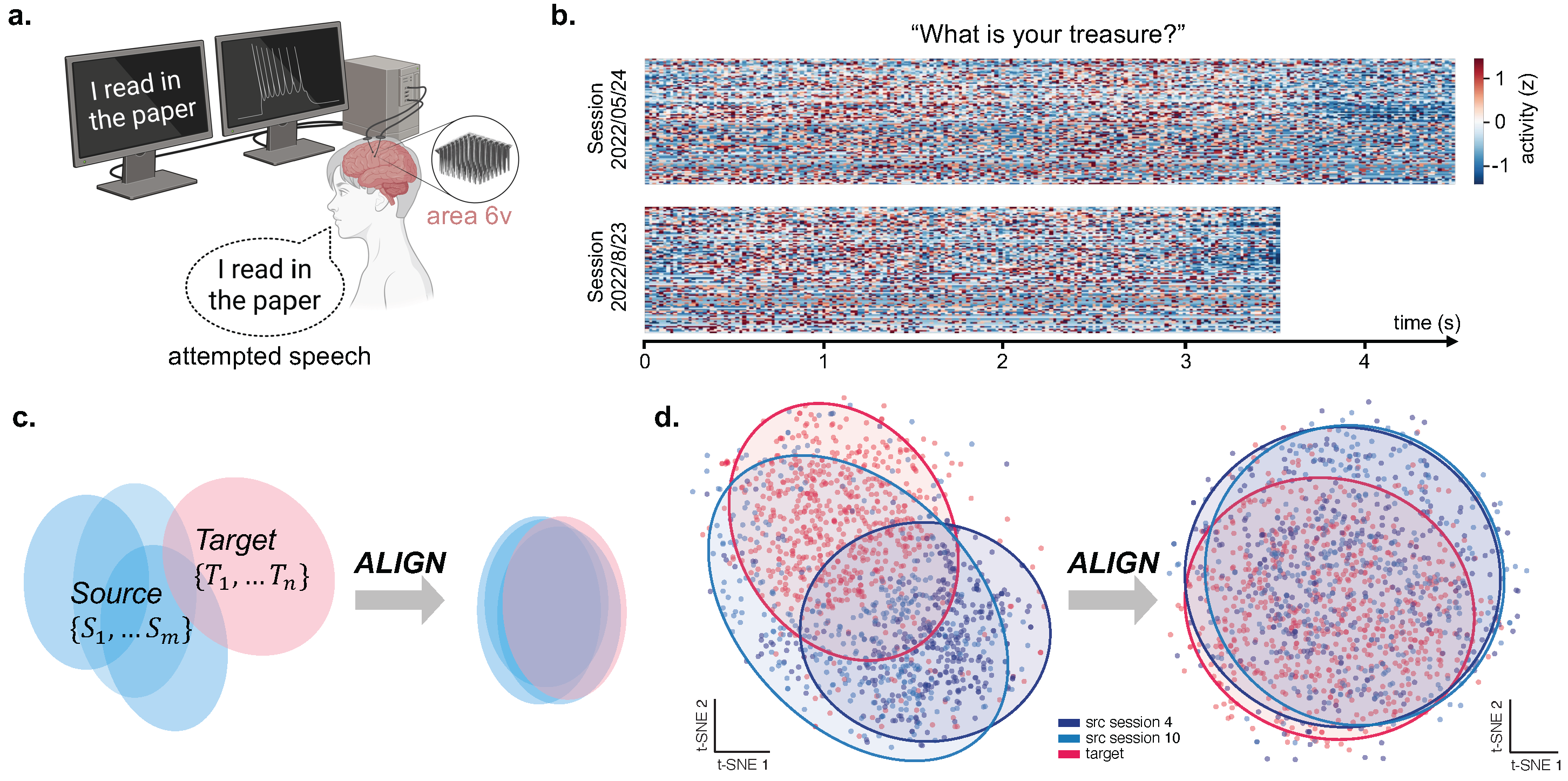}
  \caption{\textbf{Dataset and domain shift across sessions.}
    \textbf{a.} Attempted speech dataset. Intracortical neural activity was recorded from a participant with amyotrophic lateral sclerosis (ALS) while they attempted to speak sentences presented on a screen. Although vocalizations were produced, the speech was not intelligible.
    \textbf{b.} Example neural activity corresponding to the same attempted sentence recorded in two different sessions, illustrating session-dependent neural dynamics variabilities.
    \textbf{c.} Illustration of distribution shift in source sessions and target sessions caused by nonstationarities in neural dynamics and the session-invariant distributions with adaptation.
  \textbf{d.} t-SNE visualization of the latent embeddings used for phoneme decoding (input to the phoneme classifier), showing source sessions (blue) and target sessions (red) before and after ALIGN.}
  \label{fig:motivation}
\end{figure*}

In machine learning, domain adaptation aims to mitigate session-dependent distribution shift by leveraging labeled source data together with unlabeled target data, typically by learning representations that are invariant across domains. Common approaches include adversarial objectives (\cite{dann, mdan}) and importance re-weighting (\cite{mmd}). In neuroscience, related ideas have been used to align neural representations across days using adversarial training (\cite{adan, cyclegan}), improve robustness to channel/unit turnover via masking (\cite{ucla}) or permutation-invariant population models (\cite{spint}), and explicitly match latent manifolds across sessions (\cite{nonsta2, nomad}). Other work applies test-time adaptation (TTA) using pseudo-labels to update models online at deployment (\cite{ucla, corp}).

However, extending these approaches to brain-to-text decoders is substantially harder. While most of these methods are evaluated on continuous decoding tasks with dense supervision, phonemes/words are discrete categorical outputs, with no ground truth label for each neural time step because supervision is only available at the sequence level (i.e., the prompted sentence), and there is no overt behavior that can be time-aligned to neural activity.

As a result, training must rely on sequence-level objectives for alignment rather than per-frame targets, and decoder outputs are typically composed with a language model whose prior is strong but imperfect and therefore cannot reliably recover from all classes of phonetic errors. Moreover, in brain-to-text settings, test-time adaptation (TTA) via self-training can be brittle under large distribution shifts: once the phoneme error rate crosses a threshold, language-model-based pseudo-labels can become systematically wrong and errors can compound, especially when the test session is far from the last labeled training session (\cite{corp}). These challenges motivate the need for models that learn session-invariant neural representations for intracortical brain-to-text decoding and that generalize robustly to unseen sessions.

In this paper, we propose a new approach \textbf{ALIGN}, (\textbf{A}dversarial \textbf{L}earn\textbf{I}ng for \textbf{G}eneralizable Speech \textbf{N}europrostheses), for  brain-to-text decoding robust to session drifts, building on domain adaptation. Our main contributions are:
\begin{itemize}
  \item A multi-source adversarial session-invariance objective for intracortical brain-to-text decoding that aligns session representations to promote learning of session-invariant features. 
  \item An intermediate-layer adversarial regularization strategy that promotes day-invariant yet phoneme-discriminative features for decoding.
  \item Temporal Stretch Augmentation (TSA) to improve robustness to temporal nonstationarities.
\end{itemize}

\section{Related Work}

\textbf{Brain–computer interfaces (BCIs)} aim to restore communication and motor function by translating neural activity into intended actions or language (\cite{eddiechang, Moses2021, willet, ucdavis}). This is particularly important for people with severe paralysis (e.g. due to amyotrophic lateral sclerosis or brainstem stroke, for whom neuroprostheses could restore communication by converting neural signals into words. However, a major challenge for chronic BCI deployment is that neural decoding performance can degrade as recording conditions and neural representations drift over time (\cite{sussillo2016robust,heber2022drift,pun2024drift}). Models that remain stable across days can therefore reduce the need for frequent decoder recalibration (\cite{wilson2025prit}).

\textbf{Brain-to-text decoding} is a BCI setting that aims to infer a person’s intended speech directly from recorded neural activity, without direct reference to the musculoskeletal components of vocal production. \citet{willet} demonstrated strong performance with a Gated Recurrent Unit (GRU) based model to decode phonemes from neural activity, followed by beam search guided by an n-gram language model (LM) to convert phoneme to words;~\cite{ucdavis} extended this to another participant. More recent work~\cite{ucla} explores replacing the GRU with transformer decoders (e.g. relative positional embeddings, temporal patches, and masking strategies) or using pre-trained end-to-end frameworks (\cite{endtoend}).

In contrast to prior work that primarily addressed online in-session decoding and offline in-session decoding, we directly study cross-session generalization for brain-to-text decoding, for which the closest related efforts are the test-time adaptation (TTA) approaches such as those in \cite{ucla, corp}. However, TTA does not learn a generalizable embedding during training and instead relies upon post-hoc adaptation. In addition, TTA methods that rely on pseudo-labels are sensitive to pseudo-label quality: if the pseudo-label accuracy is low, TTA recalibration accuracy would also degrade (\cite{corp}). Under large session shifts, where the initial phoneme error rates are high, pseudo-labels may become systematically inaccurate, leading to runaway error compounding. These shortcomings highlight the need for methods to address the problem of drifting distributions.

\textbf{Domain adaptation} methods aim to mitigate distribution shift by combining labeled source data with unlabeled target data to learn representations that transfer across domains. Classic approaches include adversarial objectives that encourage domain-invariant features (\cite{dann, mdan}), as well as discrepancy and re-weighting methods such as Maximum Mean Discrepancy (MMD)-based alignment (\cite{mmd}). Specifically, the multi-source domain adaptation network (MDAN) framework (\cite{mdan}) extends adversarial domain adaptation to multiple source domains by coupling a task predictor with a domain classifier and aggregating per-source discrepancies to learn features that generalize to an unlabeled target domain. This multi-source perspective is especially relevant in our setting, where session identity induces substantial shifts in neural feature distributions.

However, directly porting standard domain adaptation recipes to brain-to-text is nontrivial: many alignment objectives assume per-sample labels (or low-noise pseudo-labels) in the target domain and are mostly developed for continuous-output decoding, whereas speech decoding involves discrete symbols with only sequence-level supervision. Prior BCI work has pursued cross-session robustness through adversarial alignment with autoencoder-style objectives~(\cite{adan, cyclegan}), manifold alignment of latent dynamics~(\cite{nomad}), re-weighting techniques (\cite{slicewd}), disentanglement through diffusion models (\cite{wang2024exploringbehaviorrelevantdisentangledneural}), and other approaches that align neural activity between a single source session and a single held-out target session. Beyond explicit alignment, permutation-invariant set-based encoders address changes in the recorded neural population~(\cite{spint}). Nonetheless, because most of these methods were designed for continuous labels (e.g., cursor position or EMG), they do not directly carry over to speech decoding.

Taken together, these limitations motivate a training-time approach that explicitly targets session-induced distribution shift while respecting the discrete, sequence-level supervision inherent to speech. In this work, we draw inspiration from multi-source domain adaptation frameworks like MDAN (\cite{mdan}) to treat sessions as distinct source domains and to encourage session-invariant neural representations. Next, we introduce ALIGN, which is designed to provide such a foundation and enable more reliable test-time adaptation under large session shifts.

\vspace{-0.2cm}
\section{Method}

The key intuition underlying our approach is to formulate cross-session brain-to-text decoding as a domain adaptation problem.
Given labeled neural data from multiple source sessions and unlabeled data from subsequent target sessions, our objective is to learn a session-invariant representation that preserves speech-discriminative information while reducing session-specific encoding. The goal is to improve test-time generalization without the need for retraining or fine-tuning the decoder, which is burdensome to the participant.
To this end, we introduce \textbf{ALIGN}, an adversarial multi-source single-target alignment framework built upon state-of-the-art neural decoders. ALIGN augments the base neural decoder with a domain classifier operating on latent features. Conceptually, by adversarially suppressing session-specific structure in the latent space, ALIGN retains speech-relevant information while mitigating sensitivity to session-by-session variability. We detail each component below.


\vspace{-0.2cm}
\subsection{Neural Decoder}

\begin{figure}[!htb]
  \centering
  \includegraphics[width=0.6\linewidth]{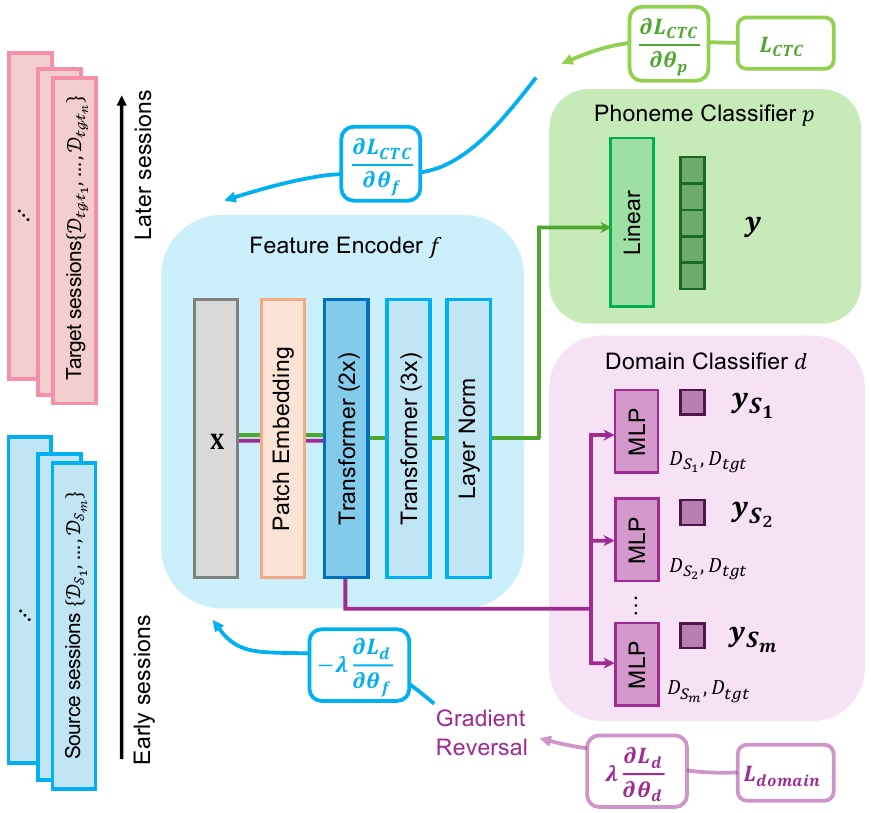}
  \caption{\textbf{ALIGN model architecture.} Our ALIGN model consists of three major modules. The feature encoder $f$ and phoneme classifier $p$ backbone shown here is instantiated with a Transformer-based decoder (\cite{ucla}). We added a domain classifier $d$, which takes the transformer encoder embedding as input and the multi-head classifiers are trained to predict whether the input is coming from source or target sessions. A gradient reversal layer is used to make the encoder learn session-invariant features.
  }
  \label{fig:method}
\end{figure}

Similar to established methods (\cite{ucla}, \cite{willet}), our neural decoder consists of two components: (1) a feature encoder $f$ and (2) a phoneme classifier $p$. We schematically illustrate the ALIGN architecture in Figure~\ref{fig:method}.
The feature encoder takes neural signals $\mathbf{X} \in \mathbb{R}^{c \times T}$ of $c$ channels and length $T$, recorded during attempted speech and produces a sequence of latent features $z_t \in \mathbb{R}^D$, a $D$-dimensional latent embedding at each time $t$, that captures the relevant information for decoding phonemes.
The phoneme classifier maps the latent sequence $\{z_t\}_{t=1}^T$ through a linear softmax layer to estimate a distribution over phonemes at each time step.


Let $y$ be the ground-truth phoneme sequence corresponding to the spoken sentence in a single trial $x \in \mathbf{X}$. We train the encoder and phoneme classifier with the connectionist temporal classification (CTC) loss (\cite{ctcloss}):
\begin{equation*}
    L_{\text{CTC}}(\theta) = -\log P_\theta(y \mid x) = -\log \sum_{\pi \mapsto y} \prod_{t=1}^T P_\theta(\pi_t \mid x),
\end{equation*}
where $P_\theta(\pi_t \vert x)$ are the probabilities of frame-level phoneme label $\pi_t$ at time $t$ given by the softmax over the encoder’s output $z_t=f(x_t; \theta_f)$, and $\theta = (\theta_f,\theta_p)$, where $\theta_f$ and $\theta_p$ denote the parameters of the encoder and phoneme classifier, respectively. The CTC loss marginalizes over all valid frame-level alignments between neural activity and the target phoneme sequence. This is essential to enabling decoding in the BCI setting where precise temporal alignment between neural signals and phoneme onsets is unavailable, since there is no overt behavior.

\subsection{Adversarial Session Invariance}

\looseness-1
Neural recordings collected across sessions exhibit distribution shifts, even though the phoneme prediction decoding task remains unchanged.
We therefore treat earlier labeled sessions as source domains and subsequent sessions as target domains, framing cross-session decoding as a domain adaptation problem. To promote generalization to target sessions and unseen test sessions, ALIGN introduces an adversarial domain classifier on the feature encoder's latent features.

Formally, the adversarial domain classifier $d(z;\theta_d)$ is constructed as a multi-head binary classifier with $m$ heads, one for each source session, where $\theta_d$ are the parameters of the domain classifier. Each head $i$ is trained to distinguish embeddings from source session $\mathcal{D}_{S_i}$ versus embeddings from all target sessions. Samples from $\mathcal{D}_{S_i}$  are routed only to head $i$, while samples from the target sessions are routed to all heads. Each head outputs a binary prediction $y_{S_i} \in \{0,1\}$ (source vs.\ target) using a three-layer multilayer perceptron (Figure~\ref{fig:method}). We train each domain classifier head with binary cross-entropy classification loss and take the mean across all $m$ classifier heads.
The overall domain loss is computed as the mean across all $m$ heads
\begin{equation}
  \mathcal{L}_{\text{domain}}(\theta)
  = \frac{1}{m}\sum_{i=1}^{m} \mathcal{L}_{\text{session}}^{(i)}(\theta).
\end{equation}
%
%

While we optimize $\theta_d$ to \emph{minimize} $L_{\text{domian}}$, we simultaneously optimize the encoder parameters $\theta_f$ to \emph{maximize} this loss with a gradient reversal layer (GRL, ~\cite{grl}). At test time, the domain classifier is inactive, and phoneme predictions are generated by the neural decoder, and then passed to the LM.
\subsection{Alternative Scheduling}

The total loss is given by
\[
  \mathcal{L}
  =
  \mathcal{L}_{\text{CTC}}
  +
  \alpha\, \lambda \, \mathcal{L}_{\text{domain}}.
\]
where $\lambda$ determines the relative importance of the domain objective in the total loss, controlling the global strength of domain alignment throughout training.

During the forward pass, the GRL acts as the identity mapping between $f$ and $d$. During backpropagation, it multiplies the gradient flowing from the domain loss into the feature encoder by $-\alpha \lambda$. Therefore, the gradient update on the encoder 
parameters become:
\[
  \frac{\partial \mathcal{L}}{\partial \theta_f}
  =
  \frac{\partial \mathcal{L}_{\text{CTC}}}{\partial \theta_f}
  -
  \alpha \lambda
  \frac{\partial \mathcal{L}_{\text{domain}}}{\partial \theta_f}.
\]

The phoneme and domain classifiers receive gradients:
\[
  \frac{\partial \mathcal{L}}{\partial \theta_p}
  =
  \frac{\partial \mathcal{L}_{\text{CTC}}}{\partial \theta_p},
  \qquad
  \frac{\partial \mathcal{L}}{\partial \theta_d}
  =
  \lambda
  \frac{\partial \mathcal{L}_{\text{domain}}}{\partial \theta_d}.
\]

We use $\alpha$ to modulate the magnitude of the reversed gradient passed at each step, governing how strongly adversarial signals influence the feature encoder. Let $s$ be the training steps, and $s_{total}$ be the total number of training steps then:
\[
  \kappa = \min\!\left(\frac{s}{s_\text{total}},\, 1\right),
\]
denotes normalized training progress. We applied a sinusoidal scheduling $\alpha(\kappa, \omega)$
\[
  \alpha(\kappa, \omega)
  =
  \frac{1}{2}
  +
  \frac{1}{2}
  \sin\!\left(\omega\,\pi\, \kappa - \frac{\pi}{2}\right).
\]

Thus, $\alpha(\kappa, \omega)$ is a sinusoidal function whose frequency is controlled by the $\omega$ parameter. This scheduling strategy allows the model to alternatively prioritize learning phoneme-discriminative structure and progressively enforcing domain invariance. Together, $\lambda$ sets the scale of domain supervision, while $\alpha(\kappa, \omega)$ dynamically regulates its effect on representation learning over the course of training.

This two-player setup eventually converges to an equilibrium where $z_t$ contains minimal session information and the encoder cannot further improve by altering $z_t$ without hurting CTC performance. Table~\ref{tab:hyperparams} lists the full set of hyperparameters.

\subsection{Temporal Stretch Augmentation}

To address variability in the duration of equivalent sequences across both repetitions and sessions observed in the data (Figure~\ref{fig:motivation}b), we developed Temporal Stretch Augmentation (TSA) . For each neural feature sequence $\mathbf{X} \in \mathbb{R}^{c \times T}$, we simulate trial-to-trial variability in the temporal dimension by generating a stretched version of the sample. Specifically, for each sample, a stretch factor $r$ is drawn from a predefined range $[r_{\min}, r_{\max}]$, and the sequence is rescaled to length
$  T' = \left\lfloor rT \right\rfloor.$
The resampled sequence $\widetilde{\mathbf{X}} \in \mathbb{R}^{c\times T'}$ is obtained by linear interpolation along the temporal axis while preserving feature dimensionality. See Appendix \ref{sec:TSA} for more details. This augmentation simulates natural variations in speech rate and neural timing without altering phoneme labels.



\subsection{Test-time Adaptation}
We used DietCORP proposed in~\cite{ucla} for test-time adaptation (TTA) with two initializations: (i) TTA from the first \textit{target} session, where we first adapt on the validation sessions stream and use the resulting weights as initialization for test sessions ; and (ii) TTA from the first \textit{test} session, where adaptation begins from the neural decoder checkpoint trained on all source sessions, at the first test trial.

Specifically, for each trial, we first feed the neural decoder's CTC logits into the 3-gram language model to obtain an LM-refined pseudo-transcription, which is then converted to a sequence of phoneme pseudo-labels. We then adapt the decoder online by minimizing the CTC loss to these pseudo-labels. For each trial, we generate $n=64$ augmented copies using additive white noise and baseline shifts and perform one gradient step, carrying the updated weights forward to subsequent trials, following the original DietCORP adaptation procedure in~\cite{ucla}.

\begin{figure*}[!htb]
  \centering
  \includegraphics[width=\linewidth]{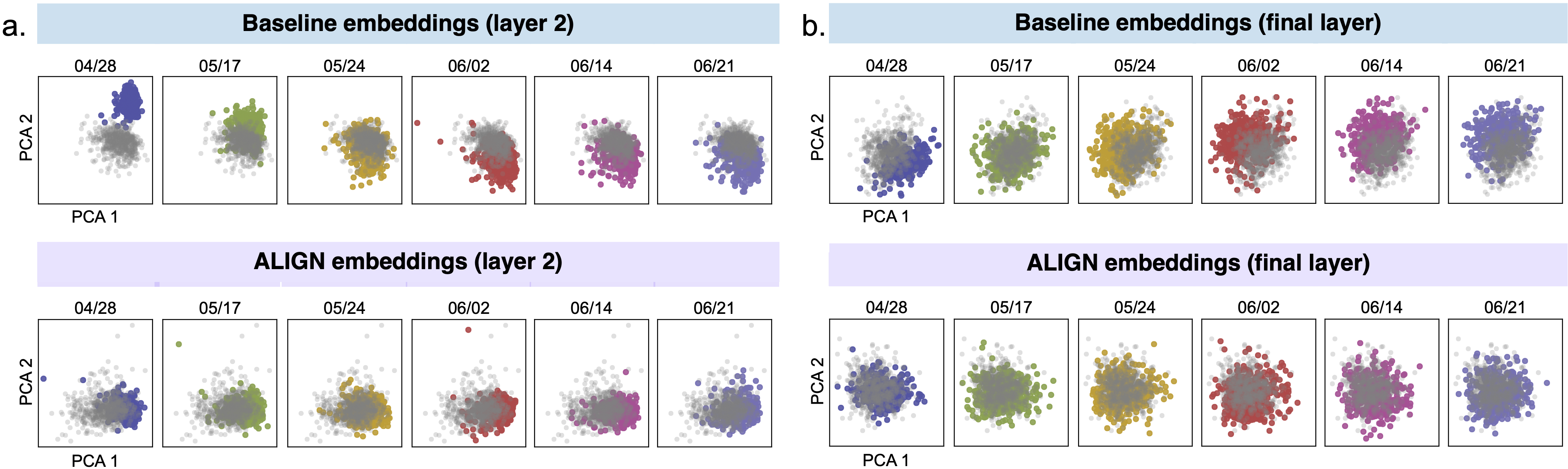}
  \caption{\textbf{Embedding visualization before and after ALIGN.}
    (\textbf{a}) PCA projections of the intermediate latent embeddings produced by the feature encoder in the transformer baseline decoder for T12 (top) and in ALIGN (bottom). Twelve source sessions are shown in color, along with three target sessions shown in gray. (\textbf{b}) Corresponding embeddings of final latent embeddings in both model.}
  \label{fig:align-pc}
\end{figure*}

\section{Experiments}
\label{sec:experiments}
\noindent \textbf{Datasets}
We evaluated ALIGN on two intracortical speech BCI datasets from participants with Amyotrophic Lateral Sclerosis (ALS), T12~(\cite{willet}), and T15~(\cite{ucdavis}). Both datasets consist of multi-session recordings from four 64-channel micro-electrode arrays (256 electrodes total) implanted in the cortical speech motor network (ventral and premotor cortex for T12; ventral premotor cortex, area 6v, area 55b, and primary motor cortex for T15). Neural activity is represented as binned features every 20 ms (256 features for T12; 512 for T15). T12 contains 10{,}850 sentences collected over 24 sessions ($\sim$4 months), and T15 contains 10{,}948 sentences collected over 45 sessions ($\sim$20 months). Each trial includes neural time-series features paired with sentence-level transcriptions and phoneme sequence labels. Full dataset details are provided in~\cite{willet, ucdavis}.


The original T12 and T15 train/validation/held-out dataset partitions are split per session (Figure~\ref{fig:dataset-split}, left). For example, in T12, the validation partition (referred to as ``test'' in~\cite{willet}) contains the last block of each day with 40 sentences; the held-out partition (referred to as ``competitionHoldOut'' in~\cite{willet}) contains the first two blocks with 80 sentences; and ``train'' contains the rest. In T15, approximately 90\% of each session's data was used for training and validation, and 10\% was used for held-out.

For cross-session generalization experiments, we restructured the train/validation/test partition as follows: 
$m$ source sessions (domains) $\{\mathcal{D}_{\text{src}_1}, \dots, \mathcal{D}_{\text{src}_m}\}$, 
$n$ subsequent unlabeled target sessions $\{\mathcal{D}_{\text{tgt}_1}, \dots, \mathcal{D}_{\text{tgt}_n}\}$ 
, and $\ell$ subsequent held-out test sessions $\{\mathcal{D}_{\text{test}_{1}},\dots,\mathcal{D}_{\text{test}_{\ell}}\}$ 
, where each session corresponds to a separate recording day (Figure~\ref{fig:dataset-split}, right). There are gap days between recording sessions (i.e., sessions are not consecutive days). For both T12 and T15, we combine the train and validation data from source days as our training set, and the train and validation data from target days as our validation set. We use the unlabeled (i.e. no phoneme labels) held-out data from all target days as the target set, and we use the combined original training and validation data from test days as our test set.

We evaluate our models on both T12 and T15 using several session-level partitions to assess cross-session generalization. For T12, we consider three partitions: (a) 12 source, 8 target, 3 test (referred to as 12--8--3), i.e. 12 sessions are used for training, 8 sessions for validation (target-domain tuning), and the remaining 3 sessions for testing; (b) 15 source, 5 target, 3 test (referred to as 15--5--3); and (c) 12 source, 4 target, 7 test (referred to as 12--4--7). The first two partitions match those of~\cite{ucla}. For T15, we consider 5 source, 13 target, 4 test. If a source session contains only training data (i.e., no validation or test partition), we use this data 
as the source data.

\subsection{Evaluation Metrics}
Phoneme error rate (PER) and word error rate (WER) are used to evaluate model decoding accuracy. PER is the number of phoneme errors (i.e., insertions, deletions, and substitutions) from all sentences over the total number of ground truth phonemes, computed directly from the output of the neural decoder after duplicated phonemes were removed. PER serves as the raw measurement of the decoding performance and validation PER was used for model selection. By feeding phoneme probabilities from the neural decoder into an n-gram language model, we can obtain the sequence of words forming a likely natural language sentence and compute the WER with respect to the ground-truth transcript of the attempted speech.

\subsection{Baselines}
\noindent \textbf{GRU-based Neural Decoder}
The GRU decoder is a recurrent neural network that maps time-varying neural activity to phoneme sequences (\cite{willet, ucdavis}). It first applies a day-specific transformation to normalize session-dependent variability in neural signals. The transformed features are then processed by a stack of gated recurrent units (GRUs), which capture temporal dependencies in the neural activity. The final recurrent representation is passed through a linear layer to produce frame-level phoneme logits, and the model is trained using CTC loss to align neural activity with phoneme sequences without requiring explicit frame-level labels. Data augmentation includes additive white noise, baseline shift, random temporal cuts, and Gaussian smoothing to improve robustness across sessions. We adopt the same augmentation scheme during training. The GRU decoder serves as baseline for both T12 and T15. 


\noindent \textbf{Transformer-based Neural Decoder}
We used the Transformer decoder of \cite{ucla} as an additional baseline for T12 only (an official checkpoint trained on T15 was not available at the time of submission). The input neural time series of $256$ channels was padded and Gaussian-smoothed, then partitioned into non-overlapping temporal patches of length $5$. Each patch ($\mathbb{R}^{256 \times 5}$) was flattened and mapped to a $m=384$-dimensional embedding with LayerNorm and a linear projection and a second LayerNorm over $m$, yielding a patch sequence that was processed by a 5-layer Transformer with causal self-attention. During training, patch-level time masking was applied. A linear projection produces phoneme logits per timestep for CTC-based phoneme decoding.

\subsection{Multi-session Generalization}
For all experiments, we performed model selection and hyperparameter tuning based on validation PER.

\noindent \textbf{T12 evaluation:} 
For participant T12, ALIGN achieved better PER across target sessions compared to the transformer baseline (Figure~\ref{fig:per_perday}), yielding an average session improvement of approximately 9\% validation PER on the 12-8-3 split. PCA projections of source and target representations of the baseline model at earlier transformer layers exhibited more noticeable session drift (Figure~\ref{fig:align-pc}a, top) than deeper layers (Figure~\ref{fig:align-pc}b, top). 
We therefore applied the adversarial invariance loss to the intermediate layers (Figure~\ref{fig:method}), which yielded the most effective drift-robust decoding performance. Applying ALIGN resulted in more spatially overlapped embeddings between each source session and the target sessions, indicating that the learned representations were less session-dependent as shown in Figure~\ref{fig:align-pc} (bottom rows) for T12 partition 12-8-3. 

\begin{figure*}[!htb]
  \centering
  \includegraphics[width=\linewidth]{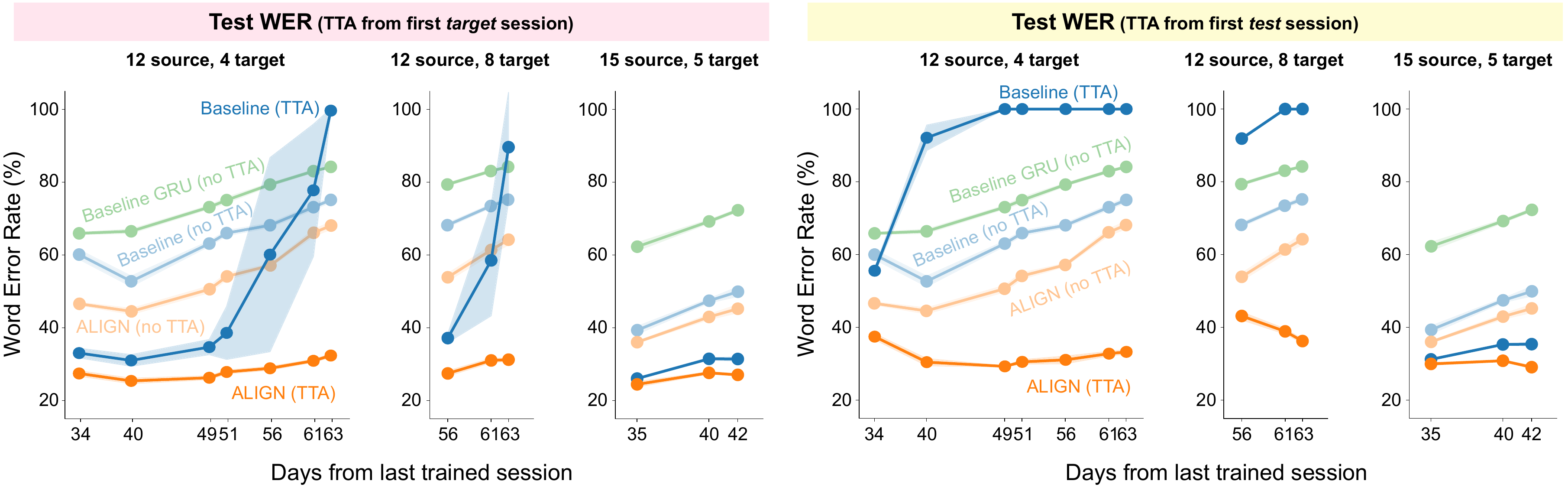}
  \caption{\textbf{ALIGN with and without TTA.}
  Test WER of GRU baseline (green) without TTA, transformer baseline (blue) and ALIGN model (orange) with and without TTA (dark and light shades), where TTA is trained from the first target session (left) or from the first test session (right). All models are tested on three different train-test partition for 5 seeds.}
  \label{fig:result-t12}
\end{figure*}

We also compared ALIGN against the baseline models for unseen test sessions using word error rate (WER). Because the transformer-based decoder in~\cite{ucla} achieves lower WER than the GRU baseline, we focus our T12 comparisons on the transformer model. We evaluated both models with and without test-time adaptation (TTA) for 5 random seeds. We evaluated three different train--test partitions and reported results with and without TTA  (Figure~\ref{fig:result-t12}). Across partitions, ALIGN consistently improved WER relative to the transformer baseline in both settings. Without TTA, ALIGN reduced test-session WER compared to both baselines (GRU: $65.93\pm 0.28 \%$; Transformer: $60.01 \pm 1.44\%$; ALIGN: $46.50 \pm 0.46\%$ on the first held-out test session in the 12--4--7 partition, with consistently lower WER on subsequent test sessions), suggesting that the adversarial objective promotes better generalization to held-out test sessions.

\noindent \textbf{Test-time adaption:} With TTA, ALIGN further improved test-session WER and, importantly, yielded greater stability across sessions than the transformer baseline with TTA (Figure~\ref{fig:result-t12}). When TTA started from the first target day, the transformer baseline remained consistently worse than ALIGN across partitions and test sessions (Figure~\ref{fig:result-t12}, left). For two of the three partitions, the transformer baseline's WER rose sharply on later test sessions, while WER of ALIGN showed around $65\%$ absolute improvement at the last test session. This pattern suggests that, under accumulating session drift, language-model pseudo-labels and self-training can become unreliable and may fail to correct, or even amplify, decoder errors, whereas ALIGN provides a more robust starting point for adaptation.

The benefit of ALIGN is even clearer when TTA started from the first \emph{test} session rather than the first target session. Because test sessions took place after target sessions, there was a comparatively larger distribution shift between the last labeled training day and the first day used for TTA, making self-training more challenging. Consistent with this, the transformer baseline with TTA initially performed moderately on the first test session $55.68 \pm 0.83\%$ but rapidly degraded (Figure~\ref{fig:result-t12}, right), reaching higher than $90\%$ WER from the second test session onward. On the other hand, WER of ALIGN with TTA remained lower, tightly bounded between $29\%$ and $33\%$ across days with a low overall WER of $32.23 \pm 0.35 \%$, without exhibiting a late-session collapse. These results suggest that ALIGN is more resilient to cross-session variability and can be effectively paired with modern TTA to maintain performance under larger session shifts.

Across the three T12 partitions, performance consistently degraded as the temporal gap between the final training session and the held-out test sessions increased. In the challenging 12--4--7 partition, only four target sessions are available for ALIGN, and the seven test sessions spaned a larger temporal distance from the last training day. The reduced number of intermediate target days limited the model's ability to gradually track session drift, resulting in larger performance drops (i.e. higher WER) for the baseline models. ALIGN mitigated this degradation by learning a more session-invariant latent representation, yielding consistently lower WER.

\noindent \textbf{T15 evaluation:} We next evaluated performance on T15, using the most challenging setup: 5 training sessions, 13 target sessions, and 3 held-out test sessions. This configuration contained the fewest training days and the largest temporal gap between the final training session and the latest test session, spanning up to 86 days. The limited labeled training data and substantial cross-session drift over time made this partition particularly demanding. ALIGN improved over the GRU baseline in test WER. When TTA was initialized from the first test session, the GRU baseline degraded on later sessions, whereas ALIGN remained lower than baseline across time (Table~\ref{tab:t15_test_wer}). 

\begin{table}[!htb]
  \centering
  \caption{T15 Test Word Error Rate (WER) (across 5 seeds)}
  \label{tab:t15_test_wer}
  \small
  \setlength{\tabcolsep}{10pt}
  \begin{tabular}{c ll}
    \cmidrule(lr){2-3}
    & \multicolumn{2}{c}{\bfseries WER (\%) (mean $\pm$ std)} \\
    \cmidrule(lr){2-3}
    \bfseries Days from last train
    & {\bfseries GRU Baseline} & {\bfseries ALIGN } \\
    \midrule
    \multicolumn{3}{c}{\textit{TTA from First Test Session}} \\
    \midrule
    71 & 51.28 $\pm$ 0.95 & 45.40 $\pm$ 0.00 \\
    84 & 86.71 $\pm$ 2.29 & 65.65 $\pm$ 0.19 \\
    86 & 99.11 $\pm$ 0.11 & 71.79 $\pm$ 0.68 \\
    \bottomrule
  \end{tabular}
\end{table}

\subsection{ALIGN model ablations}
We tested the importance of each of ALIGN's components by removing (1) adversarial loss, and (2) temporal augmentation while keeping the hyperparameter unchanged, and presented these results in the T12 12--4--7 setting in Table~\ref{tab:ablation-val-per}.


Removing the adversarial loss altogether drastically degraded validation PER from $26.30 \pm 0.52$ to $31.95 \pm 0.42 \%$. This result supports the core hypothesis of ALIGN: explicitly discouraging session-identifiable structure in the learned representation is necessary for robust cross-session generalization. Without the adversarial loss, the decoder can achieve low training error by exploiting session-specific features (e.g., electrode- or day-dependent statistics), but these cues do not transfer to new sessions, resulting in higher PER.

Second, removing temporal stretch augmentation degraded validation PER to $27.70 \pm 0.67 \%$. This indicates that TSA provides a complementary robustness beyond adversarial alignment. By expanding the temporal variability observed during training, TSA reduces reliance on narrow timing statistics of the source sessions and encourages representations that are stable under realistic temporal distortions.

\begin{table}[!htb]
  \centering
  \caption{\textbf{Ablation study on validation phoneme error rate (PER).} Models are trained with 12 source sessions and 4 target sessions. 4 seeds were ran for each condition. Validation PER are shown in mean percentage $\pm$ std.}
  \label{tab:ablation-val-per}
  \small
  \setlength{\tabcolsep}{10pt}
  \begin{tabular}{@{}l r@{}}
    \toprule
    \bfseries Ablation & \bfseries Validation PER (\%)\\
    \midrule
    ALIGN & $26.30 \pm 0.52$ \\
    w/o adversarial loss & $31.95 \pm 0.42$ \\
    w/o temporal stretch augmentation & $27.70 \pm 0.67$ \\
    \bottomrule
  \end{tabular}
\end{table}

To further examine the effect of TSA, we detached the adversarial loss and trained the transformer baseline decoder in~\cite{ucla} on (i) data without TSA and (ii) with TSA applied. To evaluate whether model trained with TSA can be more robust to temporal variability (Figure~\ref{fig:motivation}b), we computed Wasserstein distance (WD) between source--target and validation--target embedding pairs on temporally-stretched trials. The TSA-trained decoder consistently produced embeddings that remained closer to the target distribution (lower WD across stretch factors) than the same model trained without TSA (Appendix~\ref{sec:TSA}, Figure~\ref{fig:tsa-wd-analysis}). Therefore, temporal stretch augmentation likely acts as an implicit regularizer that encourages temporal invariance in the learned features, making the representation less sensitive to trial-wise speed fluctuations and reducing the mismatch between training and evaluation conditions.

\section{Discussion}
Our work addresses a practical challenge in intracortical speech BCIs: models trained on pooled data often degrade when deployed to a new recording session without labels. We propose a session-invariant learning framework that combines adversarial alignment across multiple source sessions with alignment applied at an intermediate decoder layer. This placement reduces session-discriminative structure while preserving information relevant for CTC-based phoneme decoding. Importantly, our approach operates in a few-shot, unsupervised setting on target sessions, requiring no phoneme labels during alignment. 

A second contribution is Temporal Stretch Augmentation (TSA), which generates greater temporal diversity in neural speech sequences. Biologically, animals execute motor programs over a wide range of speeds (\cite{songbird, wang2018flexibletiming, gainmodulation2018}), and biophysical properties of neurons can support downstream circuits to readout time-warped activity patterns as time-invariant representations (\cite{haim2009timewarp}). Motivated by this, TSA increases temporal variability during training by stochastically time-stretching trials. Empirically, TSA complements the adversarial loss of ALIGN and further improves performance in terms of phoneme- and word-level decoding accuracy in unseen data.

We suggest several directions for future work. Decoding performance depends jointly on the phoneme decoding model and the language model, and tighter integration or a language-model-aware training objective could further improve accuracy. Leveraging self-supervised pretraining on large corpora of unlabeled neural activity may also reduce reliance on labeled data and enhance robustness across sessions. In addition, deeper analysis of what information is removed/preserved by alignment, such as identifying which neural dimensions are most session-specific, and which phoneme errors are most impacted, could improve interpretability and guide more principled regularization. 

Overall, our findings suggest that adversarial multi-session alignment, when applied at the right representational level and paired with temporally targeted augmentation, can mitigate session-level distribution shift and advance intracortical speech decoding closer to reliable long-term deployment, and reduce the need for frequent decoder recalibration.

\vspace{0.5em}
\noindent\textbf{Acknowledgements.} We thank collaborators and colleagues for discussions and feedback, especially for the helpful conversations with Jingya Huang. Zhanqi Zhang was supported in part by the Halıcıoğlu Data Science Institute (HDSI) Ph.D. Fellowship. This work was partially supported by NSF award EFRI 2223822. This work has been made possible in part by a gift from the Chan Zuckerberg Initiative Foundation to establish the Kempner Institute for the Study of Natural and Artificial Intelligence.

\bibliography{references}

@inproceedings{
ucla,
title={Time-Masked Transformers with Lightweight Test-Time Adaptation for Neural Speech Decoding},
author={Ebrahim Feghhi and Shreyas Kaasyap and Nima Ryan Hadidi and Jonathan Kao},
booktitle={The Thirty-ninth Annual Conference on Neural Information Processing Systems},
year={2025},
}

@article{willet, title={A high-performance speech neuroprosthesis}, volume={620}, rights={2023 The Author(s)}, ISSN={1476-4687}, DOI={10.1038/s41586-023-06377-x}, abstractNote={Speech brain–computer interfaces (BCIs) have the potential to restore rapid communication to people with paralysis by decoding neural activity evoked by attempted speech into text1,2 or sound3,4. Early demonstrations, although promising, have not yet achieved accuracies sufficiently high for communication of unconstrained sentences from a large vocabulary1–7. Here we demonstrate a speech-to-text BCI that records spiking activity from intracortical microelectrode arrays. Enabled by these high-resolution recordings, our study participant—who can no longer speak intelligibly owing to amyotrophic lateral sclerosis—achieved a 9.1% word error rate on a 50-word vocabulary (2.7 times fewer errors than the previous state-of-the-art speech BCI2) and a 23.8% word error rate on a 125,000-word vocabulary (the first successful demonstration, to our knowledge, of large-vocabulary decoding). Our participant’s attempted speech was decoded  at 62 words per minute, which is 3.4 times as fast as the previous record8 and begins to approach the speed of natural conversation (160 words per minute9). Finally, we highlight two aspects of the neural code for speech that are encouraging for speech BCIs: spatially intermixed tuning to speech articulators that makes accurate decoding possible from only a small region of cortex, and a detailed articulatory representation of phonemes that persists years after paralysis. These results show a feasible path forward for restoring rapid communication to people with paralysis who can no longer speak.}, number={7976}, journal={Nature}, publisher={Nature Publishing Group}, author={Willett, Francis R. and Kunz, Erin M. and Fan, Chaofei and Avansino, Donald T. and Wilson, Guy H. and Choi, Eun Young and Kamdar, Foram and Glasser, Matthew F. and Hochberg, Leigh R. and Druckmann, Shaul and Shenoy, Krishna V. and Henderson, Jaimie M.}, year={2023}, month=aug, pages={1031–1036}, language={en} }

@misc{dann,
      title={Domain-Adversarial Training of Neural Networks}, 
      author={Yaroslav Ganin and Evgeniya Ustinova and Hana Ajakan and Pascal Germain and Hugo Larochelle and François Laviolette and Mario Marchand and Victor Lempitsky},
      year={2016},
      eprint={1505.07818},
      archivePrefix={arXiv},
      primaryClass={stat.ML},
}

@misc{mdan,
      title={Multiple Source Domain Adaptation with Adversarial Training of Neural Networks}, 
      author={Han Zhao and Shanghang Zhang and Guanhang Wu and João P. Costeira and José M. F. Moura and Geoffrey J. Gordon},
      year={2017},
      eprint={1705.09684},
      archivePrefix={arXiv},
      primaryClass={cs.LG},
}

@article{ucdavis, title={An Accurate and Rapidly Calibrating Speech Neuroprosthesis}, volume={391}, ISSN={0028-4793}, DOI={10.1056/NEJMoa2314132}, abstractNote={In a man with impaired speech from amyotrophic lateral sclerosis, an intracortical speech neuroprosthesis achieved more than 97% accuracy in decoding his intended speech and making it audible in his natural voice.}, number={7}, journal={New England Journal of Medicine}, publisher={Massachusetts Medical Society}, author={Card, Nicholas S. and Wairagkar, Maitreyee and Iacobacci, Carrina and Hou, Xianda and Singer-Clark, Tyler and Willett, Francis R. and Kunz, Erin M. and Fan, Chaofei and Nia, Maryam Vahdati and Deo, Darrel R. and Srinivasan, Aparna and Choi, Eun Young and Glasser, Matthew F. and Hochberg, Leigh R. and Henderson, Jaimie M. and Shahlaie, Kiarash and Stavisky, Sergey D. and Brandman, David M.}, year={2024}, month=aug, pages={609–618} }

@article{cyclegan, title={Using adversarial networks to extend brain computer interface decoding accuracy over time}, volume={12}, ISSN={2050-084X}, DOI={10.7554/eLife.84296}, abstractNote={Existing intracortical brain computer interfaces (iBCIs) transform neural activity into control signals capable of restoring movement to persons with paralysis. However, the accuracy of the ‘decoder’ at the heart of the iBCI typically degrades over time due to turnover of recorded neurons. To compensate, decoders can be recalibrated, but this requires the user to spend extra time and effort to provide the necessary data, then learn the new dynamics. As the recorded neurons change, one can think of the underlying movement intent signal being expressed in changing coordinates. If a mapping can be computed between the different coordinate systems, it may be possible to stabilize the original decoder’s mapping from brain to behavior without recalibration. We previously proposed a method based on Generalized Adversarial Networks (GANs), called ‘Adversarial Domain Adaptation Network’ (ADAN), which aligns the distributions of latent signals within underlying low-dimensional neural manifolds. However, we tested ADAN on only a very limited dataset. Here we propose a method based on Cycle-Consistent Adversarial Networks (Cycle-GAN), which aligns the distributions of the full-dimensional neural recordings. We tested both Cycle-GAN and ADAN on data from multiple monkeys and behaviors and compared them to a third, quite different method based on Procrustes alignment of axes provided by Factor Analysis. All three methods are unsupervised and require little data, making them practical in real life. Overall, Cycle-GAN had the best performance and was easier to train and more robust than ADAN, making it ideal for stabilizing iBCI systems over time.}, journal={eLife}, publisher={eLife Sciences Publications, Ltd}, author={Ma, Xuan and Rizzoglio, Fabio and Bodkin, Kevin L and Perreault, Eric and Miller, Lee E and Kennedy, Ann}, editor={Kemere, Caleb and Gold, Joshua I and Kemere, Caleb}, year={2023}, month=aug, pages={e84296} }

@misc{adan,
      title={Adversarial Domain Adaptation for Stable Brain-Machine Interfaces}, 
      author={Ali Farshchian and Juan A. Gallego and Joseph P. Cohen and Yoshua Bengio and Lee E. Miller and Sara A. Solla},
      year={2019},
      eprint={1810.00045},
      archivePrefix={arXiv},
      primaryClass={cs.LG},
}

@article{mmd,
author = {Borgwardt, Karsten M. and Gretton, Arthur and Rasch, Malte J. and Kriegel, Hans-Peter and Sch\"{o}lkopf, Bernhard and Smola, Alex J.},
title = {Integrating structured biological data by Kernel Maximum Mean Discrepancy},
year = {2006},
issue_date = {July 2006},
publisher = {Oxford University Press, Inc.},
address = {USA},
volume = {22},
number = {14},
issn = {1367-4803},
doi = {10.1093/bioinformatics/btl242},
journal = {Bioinformatics},
month = jul,
pages = {e49–e57}
}

@inproceedings{
spint,
title={{SPINT}: Spatial Permutation-Invariant Neural Transformer for Consistent Intracortical Motor Decoding},
author={Trung Le and Hao Fang and Jingyuan Li and Tung Nguyen and Lu Mi and Amy L Orsborn and Uygar S{\"u}mb{\"u}l and Eli Shlizerman},
booktitle={The Thirty-ninth Annual Conference on Neural Information Processing Systems},
year={2025},
}

@misc{grl,
      title={Unsupervised Domain Adaptation by Backpropagation}, 
      author={Yaroslav Ganin and Victor Lempitsky},
      year={2015},
      eprint={1409.7495},
      archivePrefix={arXiv},
      primaryClass={stat.ML},
}

@article{sussillo2016robust,
  title={Making brain--machine interfaces robust to future neural variability},
  author={Sussillo, David and Stavisky, Sergey D. and Kao, Jonathan C. and Ryu, Stephen I. and Shenoy, Krishna V.},
  journal={Nature Communications},
  year={2016},
  volume={7},
  pages={13749},
  doi={10.1038/ncomms13749}
}

@article{wilson2025prit,
  title={Long-term unsupervised recalibration of cursor-based intracortical brain--computer interfaces using a hidden Markov model},
  author={Wilson, Guy H. and others},
  journal={Nature Biomedical Engineering},
  year={2025},
  doi={10.1038/s41551-025-01536-z}
}

@article{heber2022drift,
  title={Representational drift: Emerging theories for continual learning and experimental future directions},
  author={Driscoll, Laura N. and Duncker, Lea and Harvey, Christopher D},
  journal={Current Opinion in Neurobiology},
  year={2022},
  volume={76},
  pages={102609},
  doi={10.1016/j.conb.2022.102609}
}

@article{pun2024drift,
  title   = {Measuring instability in chronic human intracortical neural recordings towards stable, long-term brain-computer interfaces},
  author  = {Pun, Tsam Kiu and Khoshnevis, Mona and Hosman, Tommy and Wilson, Guy H. and Kapitonava, Anastasia and Kamdar, Foram and Henderson, Jaimie M. and Simeral, John D. and Vargas-Irwin, Carlos E. and Harrison, Matthew T. and Hochberg, Leigh R.},
  journal = {Communications Biology},
  year    = {2024},
  volume  = {7},
  number  = {1},
  pages   = {1363},
  doi     = {10.1038/s42003-024-06784-4}
}

@inproceedings{ctcloss, 
 series={ICML ’06}, 
 title={Connectionist temporal classification: labelling unsegmented sequence data with recurrent neural networks}, 
 ISBN={978-1-59593-383-6}, 
 DOI={10.1145/1143844.1143891}, abstractNote={Many real-world sequence learning tasks require the prediction of sequences of labels from noisy, unsegmented input data. In speech recognition, for example, an acoustic signal is transcribed into words or sub-word units. Recurrent neural networks (RNNs) are powerful sequence learners that would seem well suited to such tasks. However, because they require pre-segmented training data, and post-processing to transform their outputs into label sequences, their applicability has so far been limited. This paper presents a novel method for training RNNs to label unsegmented sequences directly, thereby solving both problems. An experiment on the TIMIT speech corpus demonstrates its advantages over both a baseline HMM and a hybrid HMM-RNN.}, booktitle={Proceedings of the 23rd international conference on Machine learning}, publisher={Association for Computing Machinery}, author={Graves, Alex and Fernández, Santiago and Gomez, Faustino and Schmidhuber, Jürgen}, year={2006}, month=june, pages={369–376}, collection={ICML ’06} }

@article{nonstationary, title={Long-term stability of neural prosthetic control signals from silicon cortical arrays in rhesus macaque motor cortex}, volume={8}, ISSN={1741-2552}, DOI={10.1088/1741-2560/8/4/045005}, abstractNote={Cortically-controlled prosthetic systems aim to help disabled patients by translating neural signals from the brain into control signals for guiding prosthetic devices. Recent reports have demonstrated reasonably high levels of performance and control of computer cursors and prosthetic limbs, but to achieve true clinical viability, the long-term operation of these systems must be better understood. In particular, the quality and stability of the electrically-recorded neural signals require further characterization. Here, we quantify action potential changes and offline neural decoder performance over 382 days of recording from four intracortical arrays in three animals. Action potential amplitude decreased by 2.4% per month on average over the course of 9.4, 10.4, and 31.7 months in three animals. During most time periods, decoder performance was not well correlated with action potential amplitude (p > 0.05 for three of four arrays). In two arrays from one animal, action potential amplitude declined by an average of 37% over the first 2 months after implant. However, when using simple threshold-crossing events rather than well-isolated action potentials, no corresponding performance loss was observed during this time using an offline decoder. One of these arrays was effectively used for online prosthetic experiments over the following year. Substantial short-term variations in waveforms were quantified using a wireless system for contiguous recording in one animal, and compared within and between days for all three animals. Overall, this study suggests that action potential amplitude declines more slowly than previously supposed, and performance can be maintained over the course of multiple years when decoding from threshold-crossing events rather than isolated action potentials. This suggests that neural prosthetic systems may provide high performance over multiple years in human clinical trials.}, number={4}, journal={Journal of Neural Engineering}, author={Chestek, Cynthia A. and Gilja, Vikash and Nuyujukian, Paul and Foster, Justin D. and Fan, Joline M. and Kaufman, Matthew T. and Churchland, Mark M. and Rivera-Alvidrez, Zuley and Cunningham, John P. and Ryu, Stephen I. and Shenoy, Krishna V.}, year={2011}, month=aug, pages={045005}, language={eng} }

@article{nonsta2, title={Intra-day signal instabilities affect decoding performance in an intracortical neural interface system}, volume={10}, ISSN={1741-2552}, DOI={10.1088/1741-2560/10/3/036004}, abstractNote={OBJECTIVE: Motor neural interface systems (NIS) aim to convert neural signals into motor prosthetic or assistive device control, allowing people with paralysis to regain movement or control over their immediate environment. Effector or prosthetic control can degrade if the relationship between recorded neural signals and intended motor behavior changes. Therefore, characterizing both biological and technological sources of signal variability is important for a reliable NIS.
APPROACH: To address the frequency and causes of neural signal variability in a spike-based NIS, we analyzed within-day fluctuations in spiking activity and action potential amplitude recorded with silicon microelectrode arrays implanted in the motor cortex of three people with tetraplegia (BrainGate pilot clinical trial, IDE).
MAIN RESULTS: 84% of the recorded units showed a statistically significant change in apparent firing rate (3.8 ± 8.71 Hz or 49% of the mean rate) across several-minute epochs of tasks performed on a single session, and 74% of the units showed a significant change in spike amplitude (3.7 ± 6.5 µV or 5.5% of mean spike amplitude). 40% of the recording sessions showed a significant correlation in the occurrence of amplitude changes across electrodes, suggesting array micro-movement. Despite the relatively frequent amplitude changes, only 15% of the observed within-day rate changes originated from recording artifacts such as spike amplitude change or electrical noise, while 85% of the rate changes most likely emerged from physiological mechanisms. Computer simulations confirmed that systematic rate changes of individual neurons could produce a directional “bias” in the decoded neural cursor movements. Instability in apparent neuronal spike rates indeed yielded a directional bias in 56% of all performance assessments in participant cursor control (n = 2 participants, 108 and 20 assessments over two years), resulting in suboptimal performance in these sessions.
SIGNIFICANCE: We anticipate that signal acquisition and decoding methods that can adapt to the reported instabilities will further improve the performance of intracortically-based NISs.}, number={3}, journal={Journal of Neural Engineering}, author={Perge, János A. and Homer, Mark L. and Malik, Wasim Q. and Cash, Sydney and Eskandar, Emad and Friehs, Gerhard and Donoghue, John P. and Hochberg, Leigh R.}, year={2013}, month=june, pages={036004}, language={eng} }

@article{nonsta3, title={Intracortical recording stability in human brain-computer interface users}, volume={15}, ISSN={1741-2552}, DOI={10.1088/1741-2552/aab7a0}, abstractNote={OBJECTIVE: Intracortical brain-computer interfaces (BCIs) are being developed to assist people with motor disabilities in communicating and interacting with the world around them. This technology relies on recordings from the primary motor cortex, which may vary from day to day.
APPROACH: Here we quantify, in two long-term BCI subjects, the length of time that action potentials from the same neuron, or group of neurons, can be recorded from the motor cortex.
MAIN RESULTS: These action potentials are identified by their extracellular waveforms and may change within a single day, although some of these identified units can be identified consistently for weeks and even months. Features of the extracellular waveforms allowed us to predict whether a specific unit was more or less likely to remain stable over a prolonged period.
SIGNIFICANCE: A greater understanding of unit stability and instability can aid the development of motor BCIs, where the goal is to maintain a high level of performance despite changes in the recorded population. BCIs should be able to be operated without technician intervention for hours, and hopefully days, to provide the most benefit to the end-users of this technology.}, number={4}, journal={Journal of Neural Engineering}, author={Downey, John E. and Schwed, Nathaniel and Chase, Steven M. and Schwartz, Andrew B. and Collinger, Jennifer L.}, year={2018}, month=aug, pages={046016}, language={eng} }

@article{nomad, title={Stabilizing brain-computer interfaces through alignment of latent dynamics}, volume={16}, rights={2025 The Author(s)}, ISSN={2041-1723}, DOI={10.1038/s41467-025-59652-y}, abstractNote={Intracortical brain-computer interfaces (iBCIs) restore motor function to people with paralysis by translating brain activity into control signals for external devices. In current iBCIs, instabilities at the neural interface result in a degradation of decoding performance, which necessitates frequent supervised recalibration using new labeled data. One potential solution is to use the latent manifold structure that underlies neural population activity to facilitate a stable mapping between brain activity and behavior. Recent efforts using unsupervised approaches have improved iBCI stability using this principle; however, existing methods treat each time step as an independent sample and do not account for latent dynamics. Dynamics have been used to enable high-performance prediction of movement intention, and may also help improve stabilization. Here, we present a platform for Nonlinear Manifold Alignment with Dynamics (NoMAD), which stabilizes decoding using recurrent neural network models of dynamics. NoMAD uses unsupervised distribution alignment to update the mapping of nonstationary neural data to a consistent set of neural dynamics, thereby providing stable input to the decoder. In applications to data from monkey motor cortex collected during motor tasks, NoMAD enables accurate behavioral decoding with unparalleled stability over weeks- to months-long timescales without any supervised recalibration.}, number={1}, journal={Nature Communications}, publisher={Nature Publishing Group}, author={Karpowicz, Brianna M. and Ali, Yahia H. and Wimalasena, Lahiru N. and Sedler, Andrew R. and Keshtkaran, Mohammad Reza and Bodkin, Kevin and Ma, Xuan and Rubin, Daniel B. and Williams, Ziv M. and Cash, Sydney S. and Hochberg, Leigh R. and Miller, Lee E. and Pandarinath, Chethan}, year={2025}, month=may, pages={4662}, language={en} }

@misc{corp,
      title={Plug-and-Play Stability for Intracortical Brain-Computer Interfaces: A One-Year Demonstration of Seamless Brain-to-Text Communication}, 
      author={Chaofei Fan and Nick Hahn and Foram Kamdar and Donald Avansino and Guy H. Wilson and Leigh Hochberg and Krishna V. Shenoy and Jaimie M. Henderson and Francis R. Willett},
      year={2023},
      eprint={2311.03611},
      archivePrefix={arXiv},
      primaryClass={cs.HC},
}

@misc{endtoend,
      title={Decoding inner speech with an end-to-end brain-to-text neural interface}, 
      author={Yizi Zhang and Linyang He and Chaofei Fan and Tingkai Liu and Han Yu and Trung Le and Jingyuan Li and Scott Linderman and Lea Duncker and Francis R Willett and Nima Mesgarani and Liam Paninski},
      year={2025},
      eprint={2511.21740},
      archivePrefix={arXiv},
      primaryClass={cs.CL},
}

@incollection{wfst,title	= {Speech Recognition with Weighted Finite-State Transducers},author	= {Mehryar Mohri and Fernando C. N. Pereira and Michael Riley},year	= {2008},booktitle	= {{Handbook on Speech Processing and Speech Communication, Part E: Speech recognition}}}

@article{songbird, title={Temporal scaling of motor cortical dynamics reveals hierarchical control of vocal production}, volume={27}, rights={2024 The Author(s), under exclusive licence to Springer Nature America, Inc.}, ISSN={1546-1726}, DOI={10.1038/s41593-023-01556-5}, number={3}, journal={Nature Neuroscience}, publisher={Nature Publishing Group}, author={Banerjee, Arkarup and Chen, Feng and Druckmann, Shaul and Long, Michael A.}, year={2024}, month=mar, pages={527–535}, language={en} }

@article{haim2009timewarp,
  title   = {Time-Warp--Invariant Neuronal Processing},
  author  = {G{\"u}tig, R. and Sompolinsky, H.},
  journal = {PLOS Biology},
  year    = {2009},
  volume  = {7},
  number  = {7},
  pages   = {e1000141},
  doi     = {10.1371/journal.pbio.1000141}
}

@article{wang2018flexibletiming,
  title   = {Flexible timing by temporal scaling of cortical responses},
  author  = {Wang, Jing and Narain, Devika and Hosseini, Eghbal A. and Jazayeri, Mehrdad},
  journal = {Nature Neuroscience},
  year    = {2018},
  volume  = {21},
  number  = {1},
  pages   = {102--110},
  doi     = {10.1038/s41593-017-0028-6}
}

@article{gainmodulation2018,
  title   = {Motor primitives in space and time via targeted gain modulation in cortical networks},
  author  = {Stroud, Jake P. and Porter, Mason A. and Hennequin, Guillaume and Vogels, Tim P.},
  journal = {Nature Neuroscience},
  year    = {2018},
  volume  = {21},
  number  = {12},
  pages   = {1774--1783},
  doi     = {10.1038/s41593-018-0276-0}
}

@article{eddiechang, 
 title={A high-performance neuroprosthesis for speech decoding and avatar control}, 
 volume={620}, 
 rights={2023 The Author(s), under exclusive licence to Springer Nature Limited}, 
 ISSN={1476-4687}, 
 DOI={10.1038/s41586-023-06443-4},  number={7976}, journal={Nature}, publisher={Nature Publishing Group}, author={Metzger, Sean L. and Littlejohn, Kaylo T. and Silva, Alexander B. and Moses, David A. and Seaton, Margaret P. and Wang, Ran and Dougherty, Maximilian E. and Liu, Jessie R. and Wu, Peter and Berger, Michael A. and Zhuravleva, Inga and Tu-Chan, Adelyn and Ganguly, Karunesh and Anumanchipalli, Gopala K. and Chang, Edward F.}, year={2023}, month=aug, pages={1037–1046}, language={en} }

@article{Moses2021, title={Neuroprosthesis for Decoding Speech in a Paralyzed Person with Anarthria}, volume={385}, ISSN={1533-4406}, DOI={10.1056/NEJMoa2027540}, number={3}, journal={The New England Journal of Medicine}, author={Moses, David A. and Metzger, Sean L. and Liu, Jessie R. and Anumanchipalli, Gopala K. and Makin, Joseph G. and Sun, Pengfei F. and Chartier, Josh and Dougherty, Maximilian E. and Liu, Patricia M. and Abrams, Gary M. and Tu-Chan, Adelyn and Ganguly, Karunesh and Chang, Edward F.}, year={2021}, month=july, pages={217–227}, language={eng} }

@inproceedings{slicewd,
title={Sliced\nobreakdash-Wasserstein Importance Weighting for Robust Brain-Computer Interface Speech Decoding},
author={Noah Cowan and Scott Linderman},
booktitle={Data on the Brain {\&} Mind Findings},
year={2025},
url={https://openreview.net/forum?id=vCh4rDQuAs}
}

@misc{wang2024exploringbehaviorrelevantdisentangledneural,
      title={Exploring Behavior-Relevant and Disentangled Neural Dynamics with Generative Diffusion Models}, 
      author={Yule Wang and Chengrui Li and Weihan Li and Anqi Wu},
      year={2024},
      eprint={2410.09614},
      archivePrefix={arXiv},
      primaryClass={q-bio.NC},
      url={https://arxiv.org/abs/2410.09614}, 
}

\newpage

\onecolumn

\title{ALIGN: Adversarial Learning for Generalizable Speech Neuroprosthesis\\(Supplementary Material)}
\maketitle

\appendix
\section{Brain-to-text Benchmarking}\label{sec:T12suppbenchmark}
We evaluated on the \emph{Brain-to-Text Benchmark ’24} (T12)~\cite{willet}, 
and the \emph{Brain-to-Text Benchmark ’25} (T15)~\cite{ucdavis}.
We apply the same preprocessing pipeline as the baseline neural decoders, following the procedures described in the original benchmark releases.

\section{Phoneme Error Rate}

We selected models based on each dataset's validation phoneme error rate (PER). ALIGN consistently achieved the lowest PER across all validation and test sessions in the cross-session setting (Figure~\ref{fig:per_perday}).

For the T12 12–8–3 partition, on the validation set (8 sessions), the Transformer baseline PER increased from $24.92 \pm 0.17\%$ (5 days from last training day) to $49.21\pm0.33\%$ (51 days), with a session-average of $38.74\pm0.43\%$. ALIGN ranges from $22.07\pm0.50\%$ to $37.14\pm0.88\%$, achieving a lower session-average of $30.46\pm0.52\%$, corresponding to an absolute improvement of over 8\% points in PER. On the held-out test sessions (56--63 days away from last training), the Transformer baseline obtains a session-average PER of $52.61\pm0.24\%$, while ALIGN reduces this to $43.39\pm0.62\%$, yielding an improvement of approximately 9\% absolute PER (Figure~\ref{fig:per_perday}).

\begin{figure}[!htb]
  \centering
    \centering
    \includegraphics[width=0.6\linewidth]{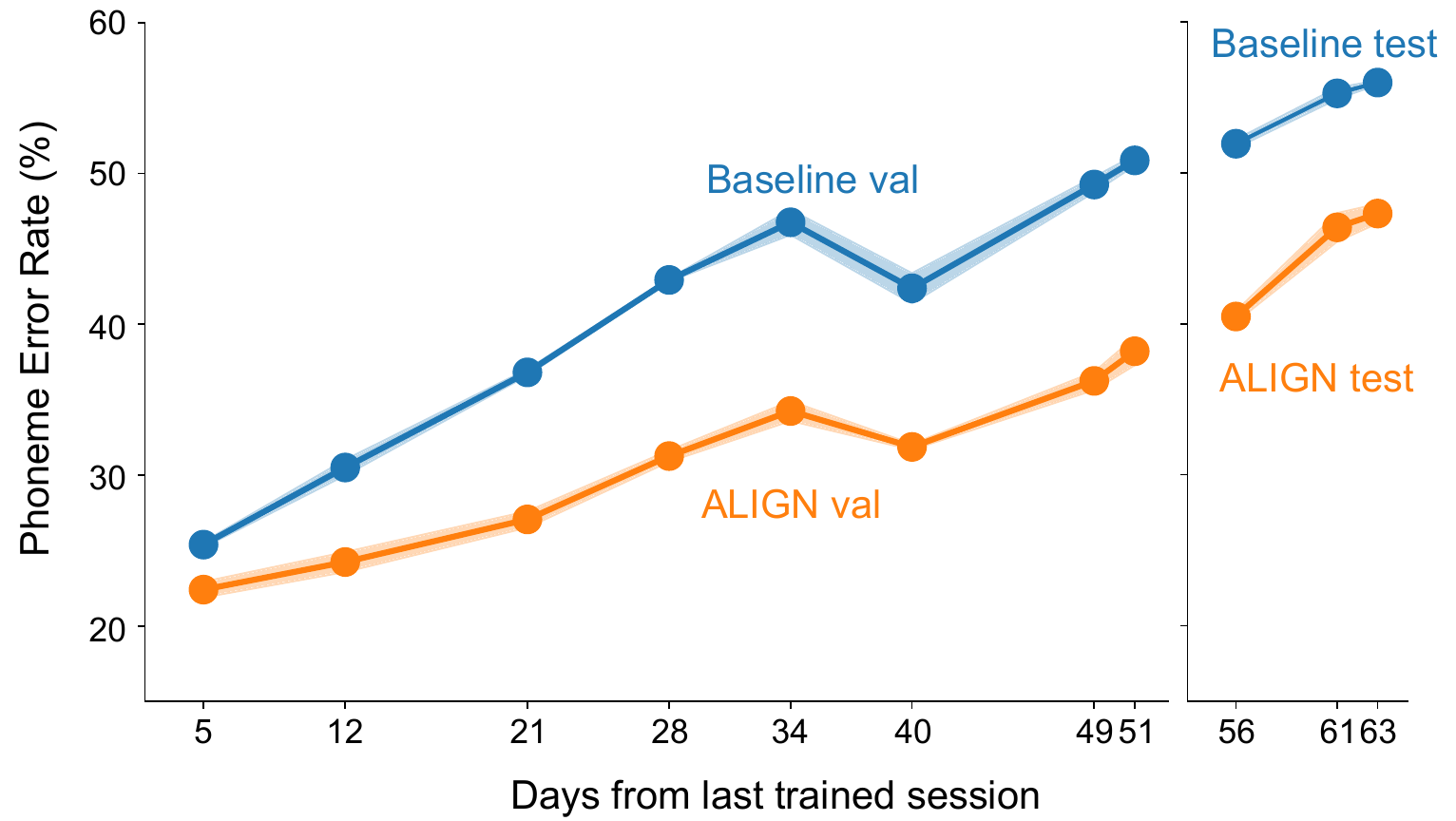}
    \caption{\textbf{Per-day phoneme error rate (PER).}
      Mean and std of PER of the transformer baseline (blue) and ALIGN (orange) on T12 12--8--3 dataset.
      The first eight sessions correspond to validation of target sessions,
    and the last three correspond to held-out test sessions.}
    \label{fig:per_perday}
\end{figure}

These results demonstrate that ALIGN improves phoneme decoding accuracy and reduces cross-session performance degradation as temporal distance from the last training session increases. The reported PER reflects raw neural decoder performance without test-time adaptation (TTA), directly measuring cross-session generalization of the learned representation. The consistently lower PER values therefore indicate improved cross-session robustness. Combined with prior findings that Transformer decoders outperform GRU baselines~\cite{ucla}, ALIGN achieves the best performance among the evaluated methods in the cross-session brain-to-text decoding setup.

\begin{figure}[!htb]
  \centering
    \centering
    \includegraphics[width=0.6\linewidth]{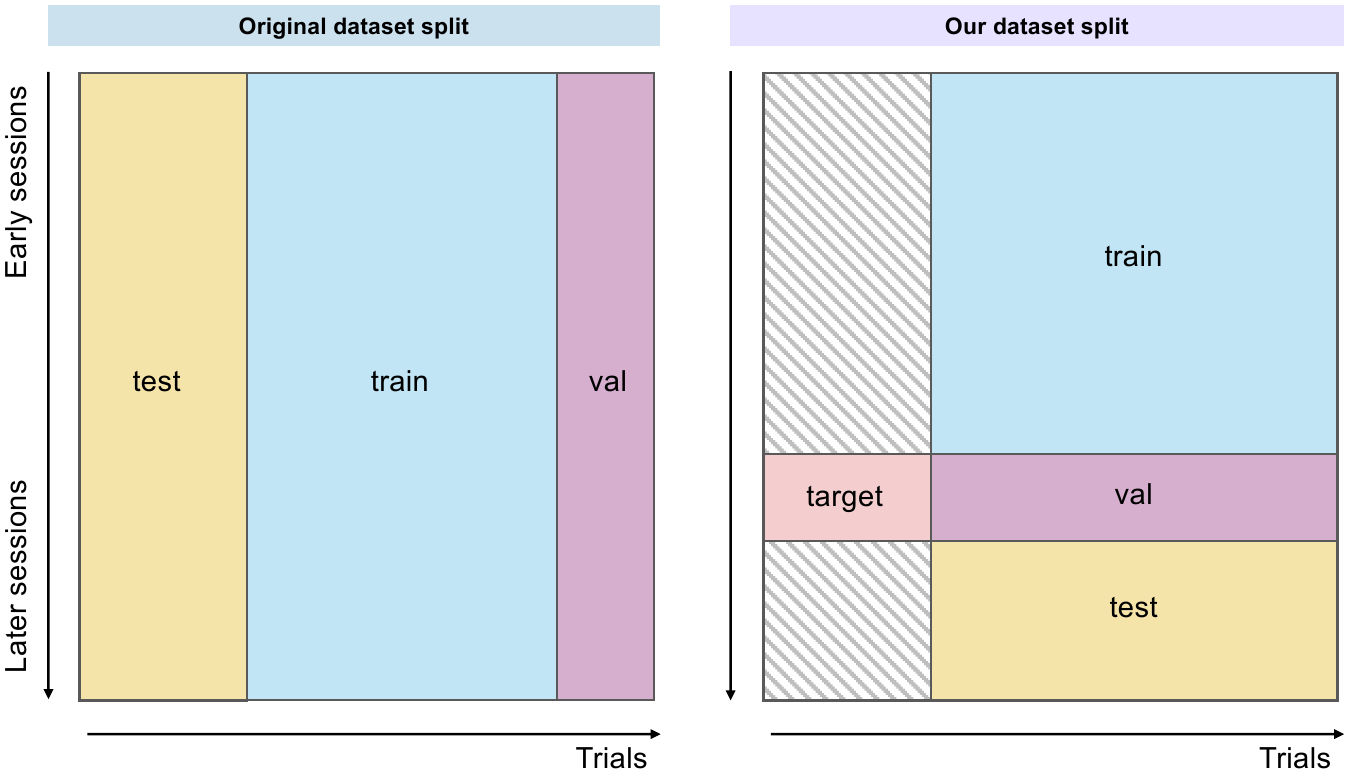}
    \caption{\textbf{Dataset Partition.}
      Left: In the official \emph{Brain-to-Text Benchmark ’24} partition, each recording day is internally divided into train, validation, and competitionHoldOut (test) blocks (except for a few sessions that lack a test block), so most sessions contribute data to every partition.
    Right: We reorganize data by session into source (train), unlabeled target (for unsupervised alignment), labeled validation (for model selection via validation PER), and fully unseen test sessions to evaluate cross-session generalization.}
    \label{fig:dataset-split}
\end{figure}

\section{ALIGN on target and test days}

\begin{figure}[!htb]
  \centering
  \includegraphics[width=\linewidth]{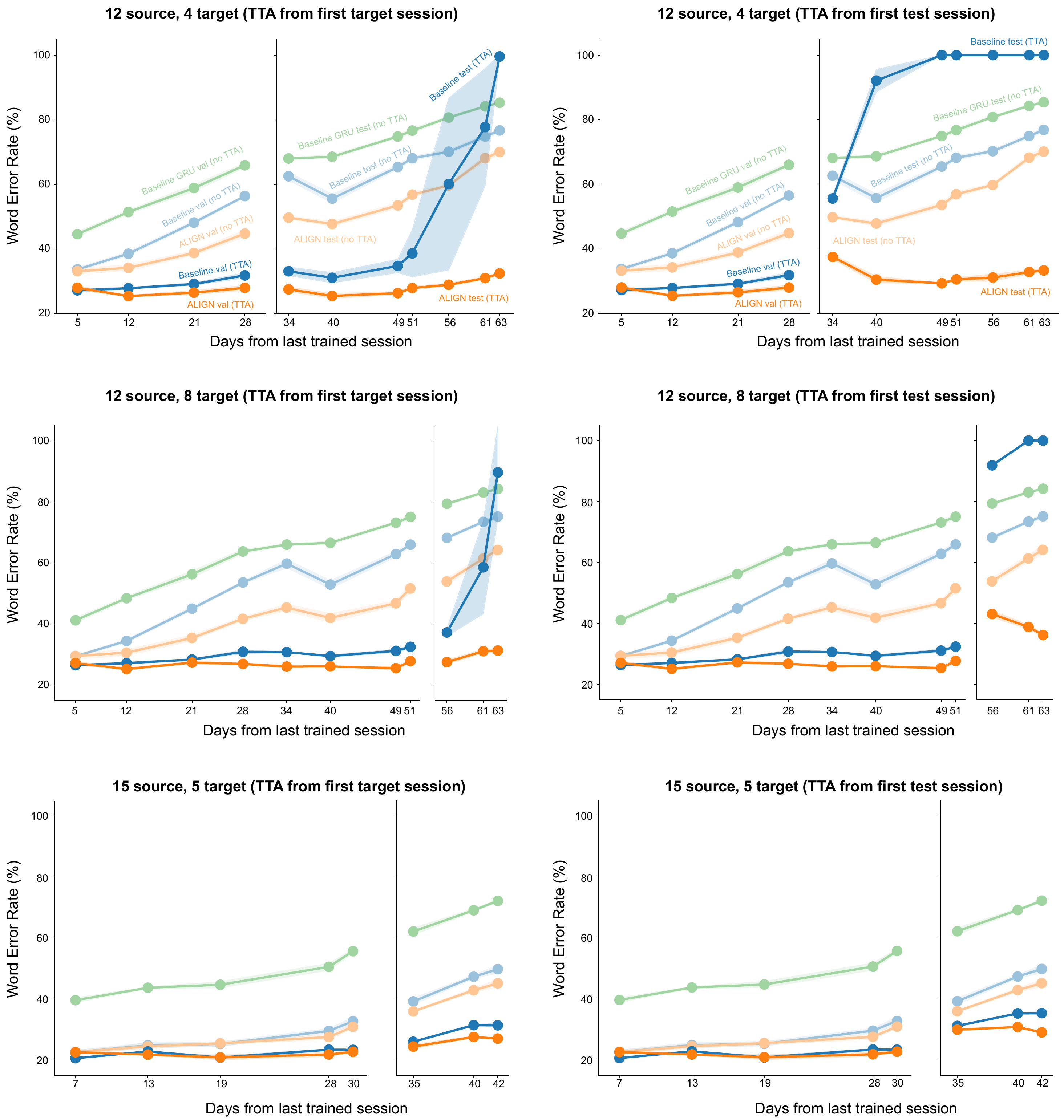}
  \caption{\textbf{ALIGN on target and test sessions.} Validation and test WER (\%) of GRU baseline (green) without TTA, transformer baseline (blue) and ALIGN model (orange) with and without TTA (dark and light shades), where TTA is trained from the first target session (left) or from the first test session (right). All models are tested on three different train-test partition for 5 seeds.}
  \label{fig:target-test-t12}
\end{figure}

ALIGN leverages the distribution of the target sessions for domain alignment, but does not use any supervised CTC objective on target data. Specifically, target session neural data are used only to reduce session-related distribution shifts between source and target domains via the alignment loss. No target transcripts are used, and the CTC decoding objective is optimized exclusively on source sessions.

Consequently, the reported Phoneme Error Rate and Word Error Rate on the target validation sessions can also be interpreted as \emph{adaptation results}, as the model is aligned to the target session distribution, but without supervised decoding on those sessions. In contrast, the unseen test sessions represent the strictest generalization and is closest to a real-world deployment setting, as neither alignment nor supervision is performed on those sessions. The model is adapted using target sessions only, and then directly evaluated on unseen test sessions.

In Figure~\ref{fig:target-test-t12}, we report the full trajectory of Word Error Rate (WER) across both target sessions and unseen test sessions, spanning up to 63 days from the last trained source day in T12.

\section{Temporal Stretch Augmentation}\label{sec:TSA}

We used temporal stretch augmentation (TSA) as a complementary approach to the adversarial structure of ALIGN to further address data variability, specifically temporal variability at the trial level. TSA is applied per trial to the neural feature sequence (time x features). The sequence is resampled along the time axis using linear interpolation with a stretch factor.

The stretch factor used in this paper is always greater than 1 due to the CTC objective of ALIGN. CTC aligns a sequence of encoder frames (length $T$) to a sequence of labels (length $L$) using blanks and repeated labels. A valid alignment exists only if the input length $T$ is at least as long as the target length $L$ (i.e. $T \geq L$) to allow the encoder have enough time steps to ``place" all labels (with optional blanks and repeats). Therefore, stretch factors less than 1 would violate such a constraint.

For each trial, a stretch factor is sampled from a specified range (e.g., $1.5-5.0\times$). During training, the data loader includes (1) the original trial and (2) a single time-stretched copy generated using a randomly sampled factor from this range.
In summary, the model is trained on a mix of original and temporally warped trials to mimic trial-to-trial temporal variability.

\begin{figure}[!htb]
  \centering
  \includegraphics[width=\linewidth]{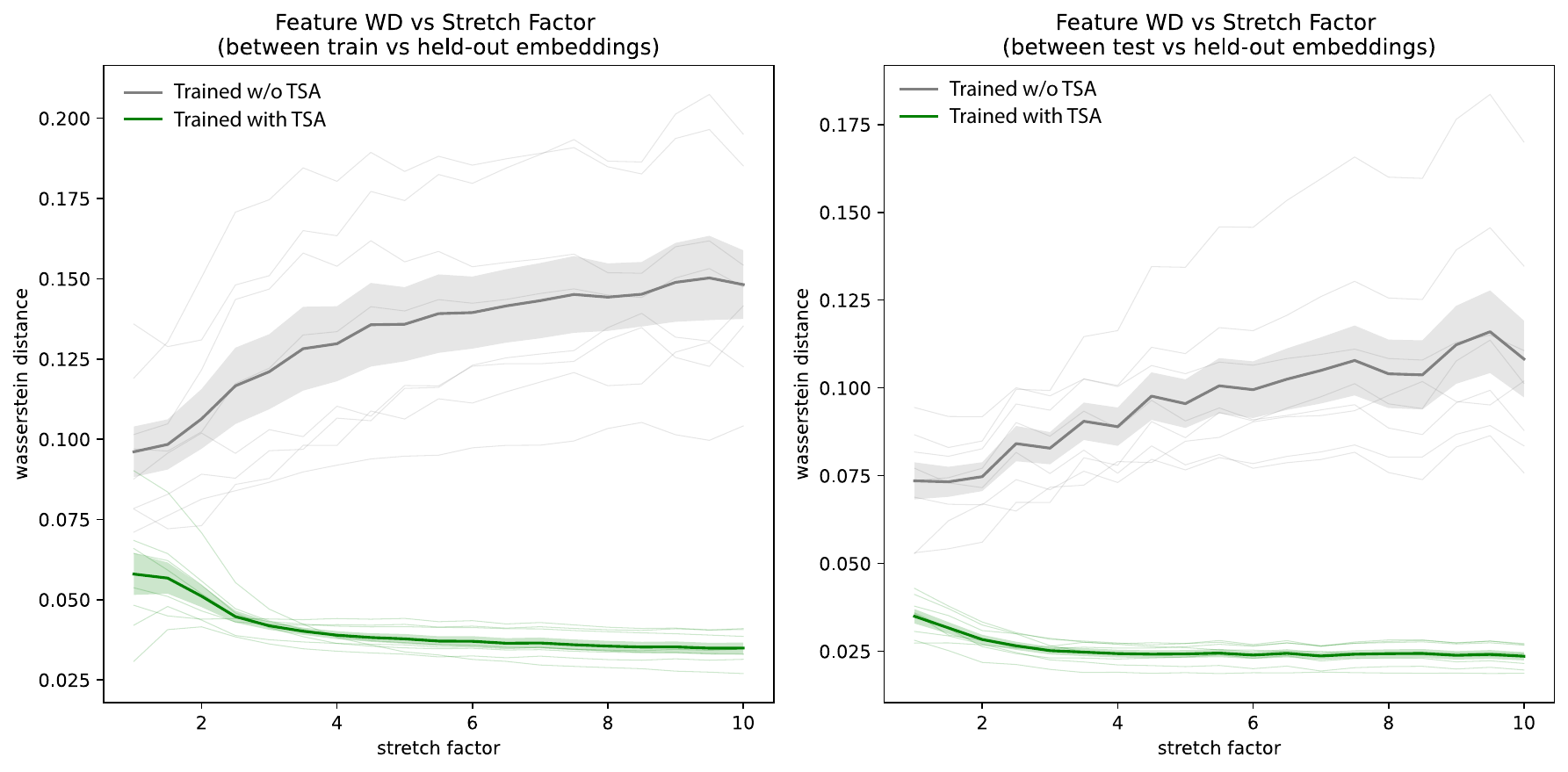}
  \caption{\textbf{Embedding analysis of TSA-trained decoder.} Feature-wise Wasserstein distance (WD) between augmented embeddings and a held-out set as a function of the stretch factor. Left: WD computed between source embeddings and target embeddings (average across the first 10 embedding dimensions); Right: WD computed between validation embeddings and target embeddings (average across the first 10 embedding dimensions). Thick lines show the mean across days and shaded bands denote $\pm$SEM; faint lines show individual day traces. Gray indicates the decoder trained on data without TSA, and green indicates decoder trained on data with TSA.}
  \label{fig:tsa-wd-analysis}
\end{figure}

\section{Embeddings of ALIGN}
\paragraph{Session-Distinct Embedding Dimension Analysis.}
To quantify session-specific variability in the learned representation, we measured the 1D Wasserstein distance between source and target distributions along each embedding dimension in the final latent space. Specifically, for each dimension $j=1,\dots,D$, we compute
\[
W_j = W\big( z^{\text{src}}_j, z^{\text{tgt}}_j \big),
\]
where $z^{\text{src}}_j$ and $z^{\text{tgt}}_j$ denote the marginal distributions of the $j$-th embedding coordinate for source and target sessions, respectively. 
We first rank dimensions in the transformer baseline from largest to smallest $W_j$, thereby identifying the most session-distinct components of the representation. 
We then compute the same distances after applying ALIGN. 
As shown in Figure~\ref{fig:w_dist_align_full}, ALIGN aligned the activation distributions across all dimensions, indicating that it suppresses session-specific variability and promotes tighter alignment between source and target distributions in the final embedding space.

\begin{figure}[!htb]
  \centering
  \includegraphics[scale=0.6]{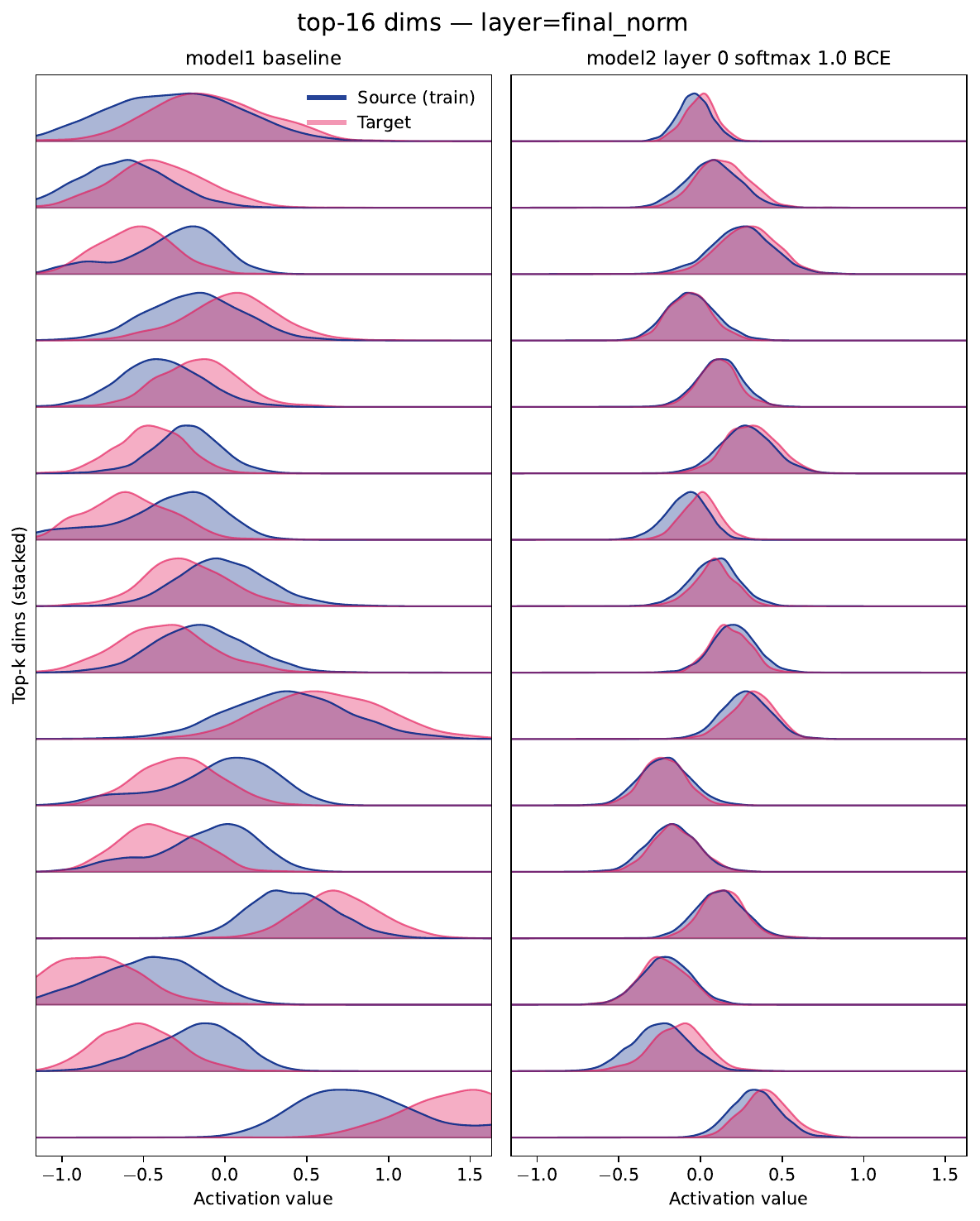}
  \caption{\textbf{ALIGN reduces the session distributional gap.} Top-$k$ most session-distinct embedding dimensions before (left) and after ALIGN (right).}
  \label{fig:w_dist_align_full}
\end{figure}


\section{Language Model}

The n-gram language model (LM) is trained on the OpenWebText2 corpus and constructed using a CMU-dictionary–based phoneme lexicon. As in~\cite{willet, ucla, ucdavis}, we use the 125{,}000-word 3-gram language model implemented in Kaldi and integrated through a Weighted Finite State Transducer (WFST) beam search decoder (\cite{wfst}). The LM is loaded into a custom Python runtime and used for phoneme-to-word decoding.

During decoding, the neural decoder produces a sequence of frame-level phoneme posterior probabilities. Beam search is applied over these probabilities to generate multiple candidate hypotheses, where each candidate beam $b$ corresponds to a complete phoneme sequence obtained after collapsing repeated labels and removing blanks.

For each candidate beam $b$, we compute a joint score by combining the encoder likelihood and the $n$-gram language model likelihood using an $\mathtt{acoustic\_scale}$:
\begin{equation}
  \text{score}(b) = \mathtt{acoustic\_scale}\, \log P_{\text{enc}}(b) + \log P_{\text{ngram}}(b).
\end{equation}
Here, $P_{\text{enc}}(b)$ denotes the probability of the phoneme sequence under the phoneme classifier (CTC), and $P_{\text{ngram}}(b)$ is the probability assigned by the $n$-gram language model to the same phoneme sequence. The beam search is performed on the composed WFST graph, and the hypothesis with the highest combined score is selected as the final decoded transcription.

To discourage excessive blank emissions from the encoder, blank probabilities are penalized by dividing them by a fixed constant during decoding.

All language model hyper-parameters matched those reported in~\cite{willet,ucla} for T12 and~\cite{ucdavis} for T15. Specifically, for the 3-gram language model on T12, we used a beam width of 18, set $\mathtt{acoustic\_scale} = 0.8$, and applied a blank penalty of $\log(2)$. For T15, we used a beam width of 17, set $\mathtt{acoustic\_scale} = 0.5$, and applied a blank penalty of $\log(9)$.

\section{Adversarial Classifier and TTA Implementation Details}
Each classifier head is a multi-layer perceptron with two hidden layers of size $d_h = 256$, each followed by ReLU activation and dropout. For test time adaptation, because DietCORP was not evaluated on T15, we implement the same DietCORP adaptation logic while adopting data augmentation hyperparameters reported in~\cite{ucdavis}. The full hyperparameters for the models are provided below.

\section{Hyperparameters}
\begin{table}[!h]
  \centering
  \caption{Transformer Decoder Training and Domain Adaptation Hyperparameters}
  \label{tab:hyperparams}
  \begin{tabular}{rl}
    \toprule
    \bfseries Parameter & \bfseries Value \\
    \midrule
    Optimizer & AdamW \\
    Learning rate & 0.001 \\
    Weight decay & $1\times10^{-5}$ \\
    Batch size & 64 \\
    Epochs & 500 \\
    Scheduler & Multistep (milestone=300) \\
    $\gamma$ & 0.1 \\
    Depth & 5 \\
    Embedding dim & 384 \\
    Heads & 6 \\
    Dropout & 0.35 \\
    Input dropout & 0.2 \\
    Patch size & (5, 256) \\
    Rep. layer index & 2 \\
    $\lambda_{\text{DANN}}$ & 0.6 \\
    $\lambda_{\text{src}}$ & 1 \\
    $\lambda_{\text{tgt}}$ & 1 \\
    Disc. hidden dim & 256 \\
    Disc. LR multiplier & 0.6 \\
    Alpha type & Alternative ($\omega=16$) \\
    Binary loss & True \\
    Target loss & True \\
    Stretch range & (1.5, 5) \\
    Max mask pct & 0.075 \\
    Num masks & 20 \\
    White noise SD & 0.2 \\
    Baseline Shift & 0.05 \\
    Gaussian smooth width & 2 \\
    Test-time Adaptation learning rate & 3e-3\\
    \bottomrule
  \end{tabular}
\end{table}

\begin{table}[!h]
  \centering
  \caption{GRU Decoder Training and Domain Adaptation Hyperparameters for T12 and T15}
  \label{tab:hyperparams-gru}
  \begin{tabular}{rll}
    \toprule
    \bfseries Parameter & \bfseries T12 & \bfseries T15 \\
    \midrule
    Optimizer & Adam & AdamW \\
    Learning rate scheduler & None & Cosine \\
    Learning rate warmup batches & -- & 1000 \\
    Training & 73 epochs & 120000 batches \\
    Batch size & 64 & 64 \\
    $\beta_0,\beta_1$ & (0.9, 0.999) & (0.9, 0.999) \\
    $\epsilon$ & 0.1 & 0.1 \\
    LR (start / end) & 0.02 / 0.02 & 0.005 / 0.0001 \\
    Weight decay (main) & $1\times10^{-5}$ & 0.001 \\
    Grad norm clip & -- & 10 \\
    Neural dim  & 256 & 512 \\
    \# GRU layers & 5 & 5 \\
    GRU hidden units & 1024 & 768 \\
    Bidirectional & False & False \\
    GRU dropout & 0.4 & 0.4 \\
    Input layer dropout & 0 & 0.2 \\
    Patch size / stride & 32 / 4 & 14 / 4 \\
    White noise std & 0.8 & 1.0 \\
    Baseline shift (constant offset std) & 0.2 & 0.2 \\
    Random cut (timesteps) & -- & 3 \\
    Smoothing kernel std & 2.0 & 2.0 \\
    \midrule
    \multicolumn{3}{l}{\bfseries ALIGN} \\
    \midrule
    Rep. layer index & -- & 2 \\
    Discriminator hidden dim & -- & 256 \\
    Discriminator LR multiplier & -- & 0.8 \\
    Discriminator weight decay & -- & $1\times10^{-5}$ \\
    Alpha type & -- & Alternative ($\omega=24$) \\
    Alpha warmup steps & -- & 5 \\
    Domain dropout & -- & 0.0 \\
    TTA learning rate & -- &  $5\times10^{-4}$ \\
    Stretch range & -- & (1.5, 5.0) \\
    \bottomrule
  \end{tabular}
\end{table}

\end{document}